\documentclass{article}%{ieeeaccess}
\usepackage{arxiv}

\usepackage[utf8]{inputenc} % allow utf-8 input
\usepackage[T1]{fontenc}    % use 8-bit T1 fonts
\usepackage{hyperref}       % hyperlinks
\usepackage{url}            % simple URL typesetting
\usepackage{booktabs}       % professional-quality tables
\usepackage{amsfonts}       % blackboard math symbols
\usepackage{nicefrac}       % compact symbols for 1/2, etc.
\usepackage{microtype}      % microtypography
\usepackage{lipsum}
\usepackage{amsmath}
\usepackage{graphicx}
\usepackage{longtable}
\usepackage{multirow}
\usepackage{pythonhighlight}%for python code

\usepackage{enumitem}
\setlist[itemize]{align=parleft,left=0pt..1em} % For reducing the indent of items inside the cell
\usepackage[numbers]{natbib}  % For numerical citations
\usepackage{subfig}

\title{\textbf{Data Leakage and Deceptive Performance: A Critical Examination of Credit Card Fraud Detection Methodologies
}}

\author{
  \href{https://orcid.org/0000-0001-5216-6019}{\includegraphics[scale=0.06]{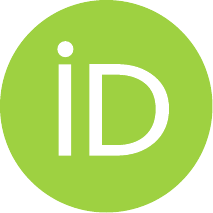}} Khizar~Hayat\thanks{Corresponding author}, 
  \\University of Nizwa,\\
  Sultanate of Oman \\
   \texttt{khizar.hayat@unizwa.edu.om},
   \AND
   \href{https://orcid.org/0000-0003-3458-0552}{\includegraphics[scale=0.06]{orcid.pdf}} Baptiste~Magnier$^{\dag}$\\%, \ddagger}$ \\
    $^{\dag}$ Euromov Digital Health in Motion, \\
    Univ Montpellier, \\
    IMT Mines Ales, Ales, France \\
    \texttt{baptiste.magnier@mines-ales.fr}\\
}

\renewcommand{\shorttitle}{A Critical Examination of Credit Card Fraud Detection Methodologies}

\begin{document}
\maketitle
\begin{abstract}\normalsize
This study critically examines the methodological rigor in credit card fraud detection research, revealing how fundamental evaluation flaws can overshadow algorithmic sophistication. Through deliberate experimentation with improper evaluation protocols, we demonstrate that even simple models can achieve deceptively impressive results when basic methodological principles are violated. Our analysis identifies four critical issues plaguing current approaches: (1) pervasive data leakage from improper preprocessing sequences, (2) intentional vagueness in methodological reporting, (3) inadequate temporal validation for transaction data, and (4) metric manipulation through recall optimization at precision's expense. We present a case study showing how a minimal neural network architecture with data leakage outperforms many sophisticated methods reported in literature, achieving 99.9\% recall despite fundamental evaluation flaws. These findings underscore that proper evaluation methodology matters more than model complexity in fraud detection research. The study serves as a cautionary example of how methodological rigor must precede architectural sophistication, with implications for improving research practices across machine learning applications.
\end{abstract}

%%%%%%%%%%%%%%%%%%%%%%%%%%%%%%%%%%%%%%%%%%%%%%%%%%%%%%%%%%%%%%%%%%%%%%%%%%%%%%%%
\section{INTRODUCTION}
The rise of deep learning has profoundly transformed how we approach complex classification and regression problems. Given its data-hungry nature, the prevailing emphasis has been on acquiring large volumes of training data to boost model performance. Meanwhile, the implementation of deep learning models has become increasingly trivialized, thanks to the proliferation of high-level APIs and libraries---especially in Python---that abstract much of the underlying complexity. Today, even large language \textcolor{red}{models (LLMs) and sophisticated neural architectures can be deployed with relatively little specialized knowledge, lowering the barrier to entry for applying advanced machine learning techniques.}

However, this convenience comes at a cost. The focus has shifted so heavily toward automation and performance metrics that critical aspects of the modeling pipeline are often overlooked or poorly documented. Researchers frequently delegate key steps to automated tools, sometimes without fully understanding or reporting them. In doing so, we risk \textit{putting the cart before the horse}---prioritizing model tuning over foundational concerns such as data quality and preparation.

The initial steps that precede model training are often loosely grouped under the term ``preprocessing,'' yet they receive scant attention in many studies. It is not uncommon for papers to devote pages to re-explaining ubiquitous evaluation metrics like accuracy, precision, recall, or confusion matrices, while providing vague or incomplete details on more foundational questions such as:

\begin{itemize}
    \item How were categorical, ordinal, or fuzzy variables handled?
    \item What strategy was used for data splitting (e.g., random, stratified, time-based)?
    \item When and how was normalization or standardization applied; and to which subsets?
    \item Were oversampling or undersampling techniques limited to the training set, or did they inadvertently affect the test data?
    \item Was feature selection or dimensionality reduction performed before or after data splitting?
\end{itemize}

This vagueness often extends to the methodological core. Sleek pipeline diagrams are commonly included but tend to omit essential details. For instance, one might encounter the use of 2D Convolutional Neural Networks (CNNs) applied to tabular data without any justification. Was there a compelling reason for employing a spatial model on non-spatial input? If so, how was the data reshaped to accommodate this architecture? Such decisions are non-trivial and require clear, transparent reporting.

These questions are far from minor. They directly affect the reproducibility, interpretability, and trustworthiness of machine learning models. When unaddressed, they may introduce silent data leakage, bias, or misleading performance metrics; ultimately undermining the credibility of the research. Thorough documentation and transparency in preprocessing are not optional; they are essential to rigorous, responsible data science.

In this work, we undertake a critical review of existing literature in light of the concerns previously discussed, using credit card fraud detection as a case study-specifically focusing on a widely referenced benchmark dataset~\cite{dalpozzolo2015}. This domain poses distinctive challenges, with extreme class imbalance being a primary issue. In such contexts, seemingly high accuracy scores are not uncommon, yet they can be misleading. Methodological oversights or missteps, whether inadvertent or otherwise, can easily lead to inflated performance metrics and a skewed perception of model efficacy.

One recurring issue we have observed is the mishandling of resampling techniques - particularly when oversampling or undersampling is performed before the train-test split - leading to data leakage. To underscore this point, we intentionally apply a simple yet flawed MLP-based approach. Despite its simplicity, the model yields impressively high metrics, thereby demonstrating how superficial performance gains can mask deeper methodological flaws.

The rest of this paper is organized as follows: Section~\ref{sec_background} provides background on credit card fraud detection and dataset characteristics. Section~\ref{sec_lit} reviews the literature on deep learning-based fraud detection. Section~\ref{sec_method} describes our methodology, followed by experimental results in Section~\ref{sec_results}. Finally, Section~\ref{sec_concl} concludes the paper.
%
% \section{Background}\label{sec_background}

% 
\section{Credit Card Fraud Detection}

The widespread use of credit and debit cards has greatly enhanced financial convenience, but it has also led to increased fraudulent activity. Payment card fraud typically involves unauthorized transactions intended to obtain goods, services, or cash. While fraud represents a tiny fraction of all transactions, the absolute financial impact is substantial~\cite{nguyen2020deep}, making it a significant area of concern for financial institutions.

\subsection{Dataset Description}
We use the popular European credit card fraud dataset~\cite{dalpozzolo2015} for benchmarking. It contains $284{,}807$ transactions made by European cardholders over two days in September 2013, out of which only $492$ are fraudulent, just $0.17\%$, highlighting extreme class imbalance.

The dataset comprises $30$ features: $28$ anonymized principal components (V1 - V28) derived via PCA, along with \texttt{Time} and \texttt{Amount}. The target variable \texttt{Class} is binary, with $1$ indicating fraud and $0$ denoting legitimate transactions. Due to privacy concerns, the original feature labels and semantics were withheld.

\begin{figure}[h!]
    \centering
    \includegraphics[width=7in]{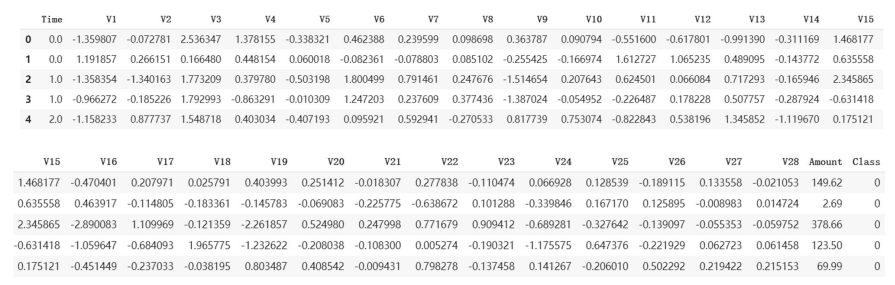}
    \caption{Snapshot of the first five rows of the dataset.}
    \label{fig:dataset_snapshot}
\end{figure}

\subsection{Handling Imbalanced Data}

In fraud detection, the misclassification of minority class instances (fraudulent cases) carries significant consequences, as failing to detect fraud can lead to financial losses and security risks. Traditional classifiers often bias toward the majority class due to the skewed distribution of data, leading to poor performance on the minority class. Resampling techniques address this issue by adjusting the class distribution, either by oversampling the minority class, undersampling the majority class, or combining both approaches to improve model performance.

\subsubsection{Oversampling Methods}
Oversampling techniques aim to increase the representation of the minority class by either duplicating existing samples or generating synthetic samples. These methods help the model learn patterns from the minority class more effectively.

\begin{itemize}
    \item \textbf{Random Oversampling}: Duplicates minority samples without introducing new information. This method is simple but can lead to overfitting due to repeated samples.
    \item \textbf{SMOTE (Synthetic Minority Oversampling Technique)}~\cite{chawla2002smote}: Generates synthetic samples between nearest neighbors of minority instances. This method helps to create a more balanced dataset and can improve model performance.
    \item \textbf{ADASYN (Adaptive Synthetic Sampling)}~\cite{he2008adasyn}: Focuses on generating synthetic samples for harder-to-classify minority samples through adaptive weighting. This method is particularly useful for improving the classification of difficult minority instances.
    \item \textbf{Borderline-SMOTE}~\cite{han2005borderline}: Creates synthetic samples near class boundaries for better discrimination. This method helps to improve the classification of minority instances near the decision boundary.
\end{itemize}

\subsubsection{Undersampling Methods}
Undersampling techniques aim to reduce the number of majority class samples to balance the class distribution. These methods help to reduce the computational complexity and improve the model's focus on the minority class.

\begin{itemize}
    \item \textbf{Random Undersampling (RUS)}: Randomly removes majority class samples. This method is simple but can lead to loss of important information.
    \item \textbf{NearMiss}~\cite{mani2003knn}: Selects majority samples based on proximity to minority instances. This method helps to retain informative majority samples.
    \item \textbf{Tomek Links}~\cite{tomek1976experiment}: Removes borderline samples to clarify decision boundaries. This method helps to improve the classification of minority instances by removing ambiguous majority samples.
    \item \textbf{Cluster Centroids}~\cite{Megha-Natarajan}: Applies K-means clustering to condense the majority class. This method helps to reduce the number of majority samples while retaining the overall distribution.
\end{itemize}

\subsubsection{Hybrid Methods}
Hybrid methods combine oversampling and undersampling techniques to balance the class distribution and improve model performance. These methods aim to leverage the strengths of both approaches.

\begin{itemize}
    \item \textbf{SMOTE-Tomek}, \textbf{SMOTE-ENN}~\cite{batista2004study}: Combine oversampling with data cleaning for improved balance. These methods help to generate synthetic minority samples and remove ambiguous majority samples.
    \item \textbf{SMOTEBoost}~\cite{chawla2003smoteboost}: Integrates SMOTE with boosting to enhance weak classifiers. This method helps to improve the performance of weak classifiers by generating synthetic minority samples.
    \item \textbf{SMOTE-SVM}~\cite{nguyen2011borderline}: Uses SVM to guide synthetic sample generation. This method helps to generate synthetic minority samples based on the decision boundary of an SVM classifier.
\end{itemize}

\subsection{Privacy Constraints and Feature Engineering}
Due to privacy-preserving transformations like PCA, meaningful original features are not available. While PCA protects sensitive data, it also results in components with uneven explanatory power, where only the top few capture substantial variance. Applying deep learning models to such transformed, sparse information may be excessive.

The literature often favors deep or complex models (CNNs, RNNs, GANs), yet we argue that focus should instead be on feature (re)engineering. By leveraging PCA components, their polynomial and pairwise combinations, and applying SMOTE for balancing, even simple models such as shallow MLPs can achieve competitive performance.

In fraud detection, minimizing false negatives (missed fraud cases) is vital. Hence, recall is a more appropriate metric than accuracy. SMOTE remains a strong choice for mitigating imbalance, and its integration with meaningful features ensures that even basic models can remain effective and interpretable.
\subsection{Performance Metrics}
% In any research project, assessing the effectiveness of machine learning algorithms is crucial. It helps identify which algorithms produce satisfactory or unsatisfactory results and evaluates their performance. 
In evaluating classification models, the confusion matrix provides essential metrics such as \textbf{accuracy}, \textbf{precision}, \textbf{recall} (sensitivity), \textbf{specificity}, and the \textbf{F1-score}. There are four key terms one can get from a confusion matrix and keeping the use case of a credit card fraud detection in perspective, these are defined as:
% \begin{figure}[h]
%     \centering
%     \includegraphics[width=5in]{logos/confusion matrix nn.png} 
%     \caption{CONFUSION MATRIX}
%     \label{}
% \end{figure}
% % 
% 
\begin{itemize}
    \item \textbf{True Positives (TP)}: Correctly predicted positive instances (e.g., fraudulent transactions correctly identified as fraud).
    \item \textbf{True Negatives (TN)}: Correctly predicted negative instances (e.g., legitimate transactions correctly identified as legitimate).
    \item \textbf{False Positives (FP)}: Incorrectly predicted positive instances (e.g., legitimate transactions flagged as fraud; Type I error).
    \item \textbf{False Negatives (FN)}: Incorrectly predicted negative instances (e.g., fraudulent transactions missed by the model; Type II error).
\end{itemize}
The key metrics and their definitions are as follows:
\[
\text{Accuracy} = \frac{\text{TP} + \text{TN}}{\text{TP+TN+FP+FN}}
\]
\[
\text{Precision} =  \frac{\text{TP}}{\text{TP+FP}}
\]
\[
\text{Recall or Sensitivity} = \frac{\text{TP}}{\text{TP+FN}}
\]
\[
\text{Specificity} = \frac{\text{TN}}{\text{TN+FP}}
\]
\[
F_1 \text{ Score}= 2 \times \frac{\text{Precision} \times \text{Recall}}{\text{Precision} + \text{Recall}}
\]

In credit card fraud detection~\cite{Aditya-Mishra}, where the cost of missing fraudulent transactions (FN) is significantly higher than false alarms (FP), recall is particularly critical to minimize undetected fraud. However, precision must also be balanced to avoid overwhelming analysts with false positives. The F1-score, which harmonizes precision and recall, is thus a key metric for assessing model performance in such imbalanced scenarios.

The \textbf{Precision-Recall Curve (PRC)} is a crucial tool for evaluating the performance of classification models, especially in imbalanced datasets where fraudulent transactions are rare. Unlike the ROC curve, which can be overly optimistic in imbalanced scenarios, the PRC provides a clearer view of the trade-off between precision and recall - two metrics that are directly relevant to fraud detection. The PRC helps in selecting an optimal threshold for the model, balancing the need to catch as many fraudulent transactions as possible (high recall) while keeping false positives manageable (high precision). This balance is critical in fraud detection, where the cost of missing fraud (false negatives) is significantly higher than the cost of false alarms (false positives).

While specificity is important for maintaining customer trust and operational efficiency, it is often secondary to recall and the $F_1$ score due to the high cost of missing fraudulent transactions. High specificity ensures that legitimate transactions are not unnecessarily flagged as fraudulent, but the primary focus remains on balancing recall and precision to minimize undetected fraud.
\section{LITERATURE REVIEW}\label{sec_lit}
% "Firstly, MLP has gained extensive usage in diverse applications due to its ability to learn complex patterns and make accurate predictions. However, its effectiveness in credit card fraud detection needs has been examined and found to be limited. The fundamental reason for this is that MLP does not possess the ability to capture temporal dependencies and sequential patterns, which are essential in detecting fraudulent activities."~\cite{Mienye2024} (watch out: first names)

Credit card fraud detection has been extensively studied using both traditional and modern machine learning approaches, with comprehensive reviews available in recent works such as~\cite{CHERIF2023,hafez2025systematic,gbadebo2023review}. While this task might appear straightforward in theory, the field has seen an overapplication of complex methods that may be unnecessarily heavy for payment card fraud detection. This trend has even led to questions about the effectiveness of certain approaches, particularly Multilayer Perceptrons (MLPs), in capturing the temporal dependencies and sequential patterns that are crucial for identifying fraudulent activities~\cite{Mienye2024}. In our view, the focus should shift from immediately applying sophisticated techniques to ensuring the correctness and efficiency of fundamental preprocessing tasks. To this end, we critically examine existing literature, identifying and analyzing potential methodological flaws in a substantial sample of previous works. 
\newcounter{mycounter}
\begin{longtable}{|p{0.03\textwidth}|p{0.25\textwidth}|p{0.25\textwidth}|p{0.075\textwidth}|p{0.075\textwidth}|p{0.075\textwidth}|p{0.09\textwidth}|}
  \caption{Machine Learning Methods } \label{tab:fulltable} \\ \hline
    \textbf{No.} & \textbf{Method/Approach} & \textbf{Flaw Identified} & \multicolumn{4}{c|}{\textbf{Reported Performance}} \\ 
    & & &\textbf{Accuracy} & \textbf{Precision} & \textbf{Recall} & \textbf{F1-Score} \\ \hline
    % Khizar for multipage table
    \endfirsthead
    \multicolumn{3}{@{}l}{\ldots continued}\\\hline
    \textbf{No.} & \textbf{Method/Approach} & \textbf{Flaw Identified} & \multicolumn{4}{c|}{\textbf{Reported Performance}} \\ 
    & & &\textbf{Accuracy} & \textbf{Precision} & \textbf{Recall} & \textbf{F1-Score} \\ \hline
    \endhead % all the lines above this will be repeated on every page
    \hline
    \multicolumn{3}{r@{}}{continued \ldots}\\
    \endfoot
    \hline
    \endlastfoot
    % end Khizar for multipage table
\stepcounter{mycounter}\arabic{mycounter} & SMOTE + ANN~\cite{sharma2021machine} & \begin{itemize} \item suspected Data leak; dataset balancing before the split \item Inconsistent results within the article \end{itemize} & 0.99 & 0.93 & 0.88 & 0.91 \\ \hline  

\stepcounter{mycounter}\arabic{mycounter} & UMAP + SMOTE + LSTM (other dimensionality reduction (UMAP etc)~\cite{benchaji2021enhanced} &  \begin{itemize} \item SMOTE on the dataset before splitting \item Dimensionality reduction on a data that is already PCA transformed\end{itemize}  & 0.967 & 0.988 & 0.919 & 0.952 \\ \hline

\stepcounter{mycounter}\arabic{mycounter} & RUS + NMS + SMOTE + DCNN ~\cite{karthika2023smart} & \begin{itemize} \item Vague on the specific order of under- and oversampling and total samples in the end. \item Precision and recall are well below 40\% - worse than a random classifier in some cases. \item No explanation of using 1D CNNs with $3\times 3$ kernels\end{itemize} & 0.972 & 0.368 & 0.392 & 0.378 \\ \hline

\stepcounter{mycounter}\arabic{mycounter} & SMOTE-ENN + boosted LSTM ~\cite{esenogho2022neural} & \begin{itemize} \item Vague on details especially data balancing. Their pseudocode suggests balancing was applied to the whole dataset.\end{itemize} & - & - & 0.996  (specificity: 0.998) & -  \\ \hline

% \stepcounter{mycounter}\arabic{mycounter} & OSCNN (SMOTE + CNN) ~\cite{abd2021deep} & totally naive; remove& 98\% & - & - & - \\ \hline

\stepcounter{mycounter}\arabic{mycounter} & SMOTE-Tomek + Bi-GRU ~\cite{sadgali2021bidirectional} & \begin{itemize} \item SMOTE-Tomek before train/test split \item BN before activation \item AUC too high par rapport the reported metrics \item readability\end{itemize} & 0.972 & 0.959 & 0.978 & 0.968  \\ \hline

\stepcounter{mycounter}\arabic{mycounter} & Borderline SMOTE + LSTM~\cite{saad2022comparative} & \begin{itemize}
    \item Improper data splitting (validation set extracted pre-oversampling) and excessive majority-class oversampling \item Misleading terminology (e.g., MLP vs. ANN) and undefined model architectures\end{itemize}& 99.9 & 80.3 & 92.1 & 85.8 \\ \hline

\stepcounter{mycounter}\arabic{mycounter} & SMOTE-Tomek + BPNN (3 hidden layers: 28+28+dropout+28)~\cite{rtayli2022efficient} &
\begin{itemize}
    \item Oversamples the entire dataset before splitting
    \item Ambiguities: sample count post-balancing, test size (25\% or 30\%?)
    \item Inconsistent metrics: AUC=1 and AUPR=0.99 incompatible with F1=0.92
\end{itemize} &
- & 0.855 & 1 & 0.922 \\\hline

\stepcounter{mycounter}\arabic{mycounter} & CAE + SMOTE~\cite{salekshahrezaee2021feature} & \begin{itemize}
    \item Claims \texttt{inverse\_transform} can reconstruct original data from PCA components alone (requires original data + PCA model)
    \item No holdout test set; relies solely on CV for final evaluation
    \item SMOTE applied globally (not per-fold), risking data leakage
    \item Multiple test evaluations may inflate performance
\end{itemize} & - & 0.920 & 0.890 & 0.905 \\\hline

\stepcounter{mycounter}\arabic{mycounter} & DAE + SMOTE + DNN (4 hidden layers: 22+15+10+5 + 2 output neurons)~\cite{Zou2019Credit} &
\begin{itemize}
    \item Bizarre DNN architecture: 4 hidden layers with aggressive shrinkage (22+15+10+5)
    \item Output layer has 2 neurons for binary classification (should be 1 + sigmoid)
    \item Unclear if normalization/standardization is applied pre-split (leakage risk)
    \item High recall at low thresholds (overfitting to SMOTE/AE) plus
    \item Sharp recall drop at threshold ~0.8 (poor probability calibration)
\end{itemize} & $0.979$ & - & $0.84$ & - \\ \hline

\stepcounter{mycounter}\arabic{mycounter} & SMOTE ahead of various ML methods with the best performance demonstrated by RF followed by an MLP with $4$ hidden layers (50+30+30+50 neurons)~\cite{Varmedja2019} & \begin{itemize}
    \item Vague methodology: lacks details on SMOTE application and feature exclusion
    \item Likely SMOTE before split (risk of data leakage)
    \item No justification for excluding 5\% of features (critical for MLP performance)
    \item Inconsistent performance: RF (F1=0.964) vs. MLP (F1=0.792) suggests overfitting
\end{itemize}& $0.999$ & $0.964$ (RF) $0.792$ (MLP) & $0.816$ & $0.884$ (RF) $0.804$ (MLP) \\ \hline

\stepcounter{mycounter}\arabic{mycounter} & CNN (Conv1D + Flatten + Dropout)~\cite{Mizher2023} &
\begin{itemize}
    \item Naive architecture: unnecessary flatten after Conv1D and excessive dropout
    \item Poor performance ($\approx 93\%$ P/R/A) vs. RF (0.99 F1)
    \item Vague oversampling details
    \item Unclear dataset reduction to 984 samples (fraud class unspecified)
\end{itemize} &
0.93 & 0.93 & 0.93 & 0.93 \\ \hline
\stepcounter{mycounter}\arabic{mycounter} & CNN (Conv1D: $32\times 2$, $64\times 2$ + Dropout + Flatten)~\cite{Ajitha2023} &
\begin{itemize}
    \item No class balancing    
    \item Misuse of Conv2D on 1D data (PCA-transformed features)
    \item Ineffective CNN use on PCA-transformed data (disrupts feature order)
    \item Forced CNN architecture (unnecessary for tabular data)
    \item Lower recall (90.24\%) despite high precision/accuracy
\end{itemize}& $0.972$ & $0.991$ & $0.902$ & $0.945$ \\ \hline

\multirow{3}{*}{\stepcounter{mycounter}\arabic{mycounter}} & a) MLP: n inputs and n neurons in each hidden layer~\cite{Ali2022} & \multirow{3}{*}{}& $0.999$ & $0.999$ & $0.999$ & $0.999$ \\\cline{2-2}\cline{4-7}

    & b) CNN: unspecified but seems to have used 2D kernels for the CNN~\cite{Ali2022} & \begin{itemize}
    \item Invalid CNN use on PCA data (no spatial structure)
    \item Misapplication of 2D kernels to 1D tabular data
    \item Lack of evaluation transparency
    \item Questionable results reliability
    \item Weak scientific justification
    \item Reported metrics (>99.9\% for ANN/CNN, 97.3\% for LSTM) seem unrealistic
\end{itemize}& $0.999$ & $0.999$ & $0.999$ & $0.999$ \\\cline{2-2}\cline{4-7}

     & c) LSTM-RNN~\cite{Ali2022} &  & $0.973$ &  $0.973$ & $0.973$ &  $0.973$ \\ \hline

\stepcounter{mycounter}\arabic{mycounter} & a) Random Oversampling (RO) + MLP: two dense hidden layers of 65 units each + 50\% dropout~\cite{Aurna2023} &  \multirow{3}{*}{}
& $0.983$ & $0.978$ & $0.987$ & $0.983$ \\\cline{2-2}\cline{4-7}

 & b) RO+CNN: Conv1D(32,2) + dropout(0.2) + BN + Conv1D(64,2) + BN + flatten+dropout(0.2) + dense(64) + dropout(0.4) + dense(1)~\cite{Aurna2023}  & 
\begin{itemize}
    % \item Misleading "federated learning" terminology
    \item CNN architecture may be suboptimal for tabular data
    \item Excessive dropout in all models (20-50\%)
    \item High accuracy without balancing masks poor other metrics
    \item Performance varies by balancing technique (SMOTE vs. random oversampling)
\end{itemize}& $0.992$ & $0.996$ &$0.987$ & $0.992$ \\\cline{2-2}\cline{4-7}

 & c) RO+LSTM: LSTM(50) + dropout(0.5) + dense(65) + dropout(0.5) + dense(1))~\cite{Aurna2023}  & & $0.957$ & $0.802$ &$0.982$ & $0.883$ \\ \hline

\stepcounter{mycounter}\arabic{mycounter} & LSTM-RNN (4x50 units)~\cite{Owolafe2021} &
\begin{itemize}
    \item Normalization before split (data leakage risk)
    \item Misconceptions about PCA re-application
    \item No class balancing (low recall: 80\%)
    \item Unnecessary implementation details
    \item High accuracy/precision (99.6\%) masks poor recall
\end{itemize} &$0.996$ & $0.996$ & $0.80$ & $0.887$ \\ \hline

\stepcounter{mycounter}\arabic{mycounter} & Time-Aware Attention RNN~\cite{Xie2023} &
\begin{itemize}
    \item No data balancing (precision: 50.07\%)
    \item High recall (99.6\%) masks poor precision
    \item Memory-intensive (performance improves with larger memory)
    \item Relies on AUC (may obscure class imbalance)
    \item Unspecified memory units
\end{itemize} &$0.958$ & $0.501$ & $0.996$ & $0.667$ \\ \hline

\stepcounter{mycounter}\arabic{mycounter} & SMOTE + AdaBoost (RF/ET/XGB/DT/LR)~\cite{ileberi2021performance} &
\begin{itemize}
    \item SMOTE before split (data leakage risk)
    \item Vague on normalization timing as well as stratification in sampling
    \item Synthetic data in the dataset may overstate performance
\end{itemize} & $0.999$ &$0.999$ & $0.999$ & $0.999$ \\ \hline

\stepcounter{mycounter}\arabic{mycounter}& SMOTE + Various Classifiers~\cite{sasank2019credit} &
\begin{itemize}
    \item Overemphasis on SMOTE+LR results
    \item Neglect of low precision (<10\%) in other methods
    \item Publisher's expression of concern
    \item Potential reliability issues
\end{itemize} & 0.970 & 0.999 & 0.970 & 0.984  \\ \hline

\stepcounter{mycounter}\arabic{mycounter} & SMOTE-Tomek + RF~\cite{mahesh2022detection} &
\begin{itemize}
    \item Ambiguous resampling order (likely wrong sequence)
    \item SMOTE-Tomek = SMOTE (no Tomek effect)
    \item Under-sampling too aggressive (492 samples)
    \item High recall focus sacrifices precision
\end{itemize} & 0.99 & 0.92 & 0.94& 0.93  \\ \hline

\stepcounter{mycounter}\arabic{mycounter} & SMOTE + XGBoost~\cite{abdulghani2021credit} &
\begin{itemize}
    \item Data leakage (preprocessing before split)
    \item Incomplete feature normalization
    \item Default classifier parameters
    \item Unrealistic perfect recall (100\%)
    \item No temporal validation
\end{itemize} &
0.999 & 0.999 & 1.00 & 0.999  \\ \hline
\end{longtable}

The study in~\cite{abdulghani2021credit} evaluates four classifiers (LR, LDA, NB, and XGBoost) using their default configurations on SMOTE-balanced data. While the authors report exceptional performance for XGBoost (accuracy: 99.969\%, precision: 99.938\%, recall: 100\%, F1: 99.969\%, AUC: 99.969\%), these results appear compromised by methodological flaws. Specifically, the preprocessing pipeline suffers from data leakage as both feature scaling/normalization and SMOTE application were performed before the train-test split. Additionally, the study's normalization approach is questionable, as it only scales the 'Time' feature without addressing the temporal nature of transaction data.

The ANN described in~\cite{sharma2021machine} has 4 hidden layers which is still fairly deep for the dataset in question. The data was balanced by applying SMOTE before doing the train/valid/test split and feeding it to the ANN, yet they got $93\%$ precision and $88\%$ recall albeit a leaky approach. The results are not consistent with those claimed in the conclusion section, later.
  
Another set of methods described in~\cite{benchaji2021enhanced} relies on swarm intelligence for feature selection, attention mechanism for classifying relevant data items, UMAP for dimensionality reduction, SMOTE for addressing data imbalance and LSTM for modeling long-term dependencies in transaction sequences. The claimed results, are again circumspect because the authors have used SMOTE on the dataset before splitting it.
  
The authors in~\cite{karthika2023smart} propose an ID Dilated Convolutional Neural Network (DCNN) for credit card fraud detection, combining SMOTE with Random Under-Sampling (RUS) and Near Miss (NM) under-sampling, though the specific order of these techniques is not explicitly stated. While their model achieves a reported accuracy of 97.27\% on the European credit card fraud dataset, this high accuracy is misleading given the dataset's extreme class imbalance (fraud prevalence of 0.172\%), where a trivial "no fraud" classifier would already achieve 99.8\% accuracy. Additionally, the paper's description of 1D CNNs with $3\times 3$ kernels and batch normalization over "feature maps" suggests a potential misapplication of 2D CNN architectures on flattened or PCA-reduced features.

Another approach~\cite{esenogho2022neural} balances the data by combining SMOTE with edited nearest neighbor (SMOTE-ENN) and then employs a strong deep learning ensemble using the long short-term memory (LSTM) neural network as the adaptive boosting (AdaBoost) technique's base learner. The authors claim to have achieved $0.996$ recall rate and a specificity of $0.998$. While everything else is described in detail, the key parts are vague, e.g. balancing before/after the dataset, normalization etc. However the data balancing algorithm outlined in the article points to the balancing of the whole dataset.

% The work reported in~\cite{abd2021deep} first tries to use a neural network or multilayer perceptron (MLP) with three ReLU based hidden layers ($30+15+5+1$ neurons) and achieve $97\%$ accuracy with $50$ epochs, after balancing the data through SMOTE. The accuracy is marginally improved ($98.9\%$ ($50$ epochs) using a CNN architecture containing three ReLU based 1D convolutional (Conv1D) layers with progressively increasing feature maps - $32$, $64$ and $64$ , respectively - along with batch normalization and dropout layers. Strangely enough, only accuracy metric is reported in the results, which make it difficult to judge the method with other criteria.
  
% The work in~\cite{du2024novel}, as the name (AE-XGB-SMOTE-CGAN) suggests, is a complex melange of autoencoder with probabilistic XGBoost based on SMOTE and Conditional Generative Adversarial Network (CGAN) For addressing class imbalance, SMOTE is complemented by the GAN.
  
Another RNN based method~\cite{sadgali2021bidirectional} uses the SMOTE-Tomek technique to address the data imbalance and, after splitting the data,  employs a Bidirectional Gated Recurrent Unit (Bi-GRU) for classification. The approach is naive, not only for applying SMOTE-Tomek before the train/test split but also for using BN before the activation function, inside the RNN module. The article also has redability issues. The reported metrics (Accuracy = $97.16\%$
Precision = $95.98\%$, Recall = $97.82\%$) seem inconsistent with the reported AUC of $99.66\%$ (unbelievably high;should be lower).
  
% AUC value increased from 0.765 to 0.929 in Chen et al. prediction model which uses k-means SMOTE and BP neural network. They achieve the highest AUC value with their model when compared to KNN logistics, SVM, random forest and tree classifiers. SVM runs noticeably less efficiently even though it displays somewhat greater accuracy~\cite{chen2021research}.
  % 
The study in~\cite{saad2022comparative} exhibits methodological flaws, particularly in data balancing and train/validation/test splits. The authors oversample the majority class but extract the validation set before oversampling, then draw test cases from the oversampled majority class, likely inflating performance metrics. 
Additionally, their terminology is inconsistent: one model is correctly labeled as an MLP (32-16-8 neurons), while another is misleadingly called an ANN (50 neurons per hidden layer, unspecified depth). Their top-performing model - a computationally heavy LSTM—relies on borderline SMOTE, casting doubt on the robustness of their results.

A three step deep learning technique~\cite{smiti2020bankruptcy} balances the data with borderline SMOTE and then uses a stacked autoencoder to extract key information for SoftMax based classification. They demonstarate a commendable AUC score but the method is still complex.
  
The authors of~\cite{rtayli2022efficient} combine SMOTE and Tomek Links to preprocess data before training a Back Propagation Neural Network (BPNN) with 3–6 hidden layers (each containing 28 neurons) and dropout regularization, particularly after the second hidden layer. However, the study suffers from ambiguities—such as the number of samples post-balancing and the test set fraction (25\% or 30\%) - as well as inconsistencies in reported metrics. Additionally, the approach risks data leakage due to balancing the entire dataset before splitting.

The study in~\cite{salekshahrezaee2021feature} compares PCA and Convolutional Autoencoder (CAE) for feature extraction, followed by SMOTE and a Random Forest classifier, recommending CAE+SMOTE with an F1-score of 90.5\%. However, their claim that \texttt{inverse\_transform} can reconstruct original transactions from PCA components alone is incorrect, as it requires the original data and PCA model. Additionally, their evaluation strategy lacks a holdout test set, risks data leakage from improper SMOTE application, and may inflate performance due to repeated testing on the same holdout set. 
% It should have contained a holdout set for final evaluation, per-fold preprocessing during CV, and reporting mean $\$ standard deviation of metrics.

The work in~\cite{Zou2019Credit} employs a denoising autoencoder (DAE) for data cleaning and SMOTE-based balancing, followed by a deep fully connected neural network (4 hidden layers: 22+15+10+5) with a 2-neuron output layer for classification. Two variants are tested: (1) without SMOTE/autoencoder (AE) and (2) with SMOTE+AE. The baseline model (w/o SMOTE/AE) achieves 97.9\% accuracy but near-zero recall, suggesting it fails to detect fraud cases. The SMOTE+AE model improves recall to 90.5\% (at the expense of accuracy) but exhibits poor probability calibration (sharp recall drop at high thresholds) and overfitting risks from synthetic data.

% The effectiveness of convolutional neural networks (CNNs) and deep neural networks (DNNs) is investigated by Joloudari et al.with  an accuracy of 99. 08\%. The hybrid SMOTE-NORM-CNN model outperforms the competing models concerning the performance~\cite{joloudari2023effective}.
% SMOTE was used by Alkhatib et al., they included input hidden and output layers in their seven-layer deep neural network design. Their model outperformed earlier efforts through evaluation using a variety of metrics attaining a 99.1\% area under the ROC curve~\cite{alkhatib2021credit}.

% Khizar
The work presented in~\cite{Varmedja2019} utilizes SMOTE prior to applying various machine learning techniques, with the best results achieved by a Random Forest classifier followed by an MLP containing four hidden layers (50+30+30+50 neurons). While the Random Forest model attains an F1-score of 0.964, the MLP's performance is notably inferior (F1=0.792), potentially indicating overfitting or inadequate training. The methodology exhibits significant weaknesses: potential data leakage from performing the train-test split after SMOTE application, unjustified exclusion of 5\% of features which might have enhanced the MLP's performance, and inconsistent results where the performance gap between RF and MLP could indicate either insufficient MLP training or feature mismatch.

The CNN-based approach in~\cite{Mizher2023} appears methodologically flawed due to its naive architecture, which includes an unnecessary flatten layer following Conv1D layers and excessive dropout layers. This overly complex model with redundant components demonstrates inferior performance ($\approx 93\%$ precision/recall/accuracy) compared to their other methods evaluated (with RF achieving 0.99 F1-score and accuracy). The study also lacks clarity regarding the oversampling process and fails to specify how the dataset was reduced to 984 samples,and how many were the fraud cases.

In~\cite{Ajitha2023}, without bothering about balancing the dataset, the authors use a CNN with two ReLU based conv2D layers (kernel sizes $32 \times 2$ and $64 \times 2$, respectively), with dropout layers and eventually a flatten layer ($64 \times 1$). Probably, the use of filters with size $k \times 2$ maybe an attempt to capture relationships between adjacent PCA features, but this should not be that useful since PCA transformation disrupts natural feature order. The use of CNNs here seems forced at best rather than necessary. The reported results (Accuracy: $97.15\%$, Precision: $99.1\%$, Recall: $90.24\%$) boasts higher performance as compared to DT, logistic regression, kNN, RF and especially XGBoost. The recall is still on the lower side, though.

The study in~\cite{Ali2022} evaluates three deep learning approaches combined with SMOTE: ANN, CNN, and LSTM RNN. While the results suggest that CNN and ANN outperform LSTM (with reported accuracy/precision/recall >99.9\% for ANN/CNN and 97.3\% for LSTM), the methodology suffers from several critical flaws. The application of CNN to PCA-transformed data is fundamentally invalid due to the lack of spatial structure in such data. Moreover, the use of 2D kernels on 1D tabular data represents a misapplication of convolutional techniques. The study lacks transparency in evaluation methodology and presents highly questionable results with weak scientific justification for the core methods employed.

The study in~\cite{Aurna2023} evaluates three deep learning architectures (CNN, MLP, and LSTM) independently, though the authors misleadingly refer to this as "federated learning." The models are trained on data balanced using various undersampling and oversampling techniques. The CNN architecture consists of a Conv1D(32, 2) layer with 20\% dropout and batch normalization, followed by a Conv1D(64, 2) layer with batch normalization, a flatten layer, 20\% dropout, a 64-unit dense layer with 40\% dropout, and a final output layer. The MLP features two dense hidden layers (65 units each) with 50\% dropout, while the LSTM comprises one LSTM layer (50 units) with 50\% dropout followed by a 65-unit dense layer with 50\% dropout before the output. As expected, the models achieve high accuracy without balancing, but other metrics suffer. Performance improves with balancing: LSTM and CNN benefit most from random oversampling, while MLP performs better with SMOTE.

The study in~\cite{Owolafe2021} implements a deep recurrent neural network with 4 stacked LSTM layers (each containing 50 hidden units), achieving high accuracy (99.6\%) and precision (99.6\%) but suffering from low recall (80\%). The methodology exhibits several flaws: normalization before train-test splitting (risking data leakage), misconceptions about PCA re-application, lack of class balancing, and inclusion of unnecessary implementation details that obscure the core methodology. While the model demonstrates strong precision, its poor recall indicates a significant bias toward the majority class, likely due to the imbalanced dataset and absence of resampling techniques.

The obsession with using RNN can be observed in~\cite{FOROUGH2021} in the form of what the authors call an "ensemble" of LSTM/GRU as RNNs. The latter are employed for classification followed by aggregation via a feed forward neural network (FFNN) as the voting mechanism. Their results on the European Cardholders Dataset (ECD) maybe good but yet it's too much complex. Same is the case with~\cite{FANAI2023} that employs AE for preprocessing, DL for classification, and a Bayesian algorithm to optimize the hyperparameters. Such overuse of deep learning methods can be found in about 11 different techniques reported in~\cite{Alarfaj2022} and strangely attributed to the adversarial approach~\cite{cartella2021}, which maybe a misreporting and seems far fetched. Similarly a CNN method is wrongly attributed to~\cite{Arora2020} which in fact has an unspecified MLP as one of the methods and that too perform very poorly on our dataset of interest with $61.4\%$ precision, $38.5\%$ sensitivity and $47.3\%$ F1 score; albeit a better specificity of $93.2\%$.

The study in~\cite{Xie2023} proposes a time-aware attention mechanism for RNNs, designed to capture users' current transactional behavior in relation to their historical patterns. While the approach aims to model behavioral periodicity effectively, it forgoes data balancing, resulting in poor precision (50.07\%) despite high recall (99.6\%). Performance improves with increased memory size (units unspecified), suggesting memory-intensive requirements. The evaluation relies on AUC rather than balanced metrics, which may obscure class imbalance issues.

The work presented in~\cite{ileberi2021performance} examines six classification algorithms (SVM, LR, RF, XGBoost, DT, and ET) both in their standard form and with AdaBoost enhancement, employing SMOTE to address class imbalance. The approach shows substantial performance gains when using AdaBoost, most notably in recall metrics (with DT recall improving from 75.57\% to 99\%), and achieves 99.95\% accuracy with RF-AdaBoost. However, the study presents several methodological limitations, 1) data leakage risk: applying SMOTE before the train-test split may contaminate the test set with synthetic samples and 2) preprocessing ambiguity: unclear specification of when normalization occurs relative to the data splitting process

The comparative analysis presented in~\cite{sasank2019credit} exhibits fundamental methodological flaws, most notably an overemphasis on the SMOTE + Logistic Regression results while neglecting to investigate the abnormally low precision (<10\%) observed in other methods. The study's credibility is further compromised by the publisher's expression of concern over the language and readability.

The study in~\cite{mahesh2022detection} evaluates three sampling techniques (under-sampling, SMOTE, and SMOTE-Tomek) combined with four classifiers (KNN, Logistic Regression, Random Forest, and SVM), though the methodology raises several concerns. First, there's ambiguity regarding the order of resampling and train/test splitting, which likely occurs in the wrong sequence. More critically, both SMOTE and SMOTE-Tomek produce identical sample counts (227,845 per class), an unusual outcome suggesting either Tomek link misconfiguration or ineffectiveness due to PCA-transformed features. Additionally, the aggressive under-sampling to just 492 samples per class risks underfitting despite appearing to yield good metrics. Finally, the evaluation's focus on recall over precision may lead to high false positive rates, as evidenced by Logistic Regression's 0.52 precision despite 0.92 recall when using SMOTE. While Random Forest achieves the highest F1-scores (0.94 with under-sampling, 0.92 with SMOTE, and 0.93 with SMOTE-Tomek), these methodological concerns undermine the reliability of the reported performance improvements.

Based on our comprehensive review of credit card fraud detection methodologies, we have identified several persistent flaws that significantly undermine the reliability and practical applicability of many studies:

\begin{itemize}
    \item Data Leakage in Preprocessing: Numerous studies perform critical preprocessing steps (normalization, SMOTE etc.) before train-test splitting, artificially inflating performance metrics through information leakage. 
    \item Intentional Vagueness in Methodology: Many works deliberately omit crucial implementation details, making replication difficult and raising questions about result validity. This includes unspecified parameter settings, ambiguous preprocessing sequences, silence about stratified sampling, and unexplained architectural choices.
    \item Inadequate Temporal Validation: Most approaches fail to account for the time-dependent nature of transaction data, neglecting temporal splitting which is essential for real-world deployment.
    \item Unjustified Method Complexity: There's a tendency to apply unnecessarily sophisticated techniques without first ensuring proper data preparation and validation, often obscuring fundamental methodological flaws.
    \item Overemphasis on Recall: Many works prioritize recall metrics at the expense of precision, leading to models with high false positive rates that would be impractical in production environments.
\end{itemize}

These persistent issues - particularly the concerning trend of intentional vagueness - highlight the urgent need for more rigorous evaluation protocols and complete methodological transparency in fraud detection research. Addressing these common pitfalls, especially the lack of full disclosure in implementation details, would significantly improve both the validity and practical utility of future studies in this domain.
% This section provides an overview of related works that used deep learning techniques for credit card fraud detection....?  with a focus on SMOTE and its variations which acts as a key factor in addressing class imbalance problems frequently encountered in datasets.
% 1. **Resampling applied before cross-validation**: If SMOTE-ENN was used on the entire dataset before splitting into training/test folds, synthetic samples could leak into the test set.
% 2. **Temporal leakage**: If transactions are time-series and not split chronologically, future data might be included in training.
% 3. **Feature engineering leakage**: If any feature engineering steps used information from the entire dataset (including test data), that would leak information.-
% Another LSTM based method~\cite{ROSELINE2022} claim to have improved performance via an attention mechanism but the claim seems to have no foundations??
% BiLSTM and BiGRU with MaxPooling layers~\cite{Hassan2020}. 
% 
\section{Our Flawed Methodology}\label{sec_method}
% 
% \begin{figure}[h!]
%     \centering
%     \includegraphics[width=7in]{logos/Methodology.png} 
%     \caption{Methodology}
%     \label{fig:Methodology}
% \end{figure} 
% % 
% The methodology used in the investigation to differentiate deceptive transactions from genuine ones is illustrated in Figure~\ref{fig:Methodology} that describes the integration of three modules central to our detection of the fraudulent card transactions, \textit{viz.} SMOTE, Polynomial Features and Artificial Neural Network (ANN). 
% Our analysis reveals that many sophisticated fraud detection methods may be unnecessarily complex when fundamental flaws persist in the evaluation pipeline. The most critical issues we've identified include: (1) pervasive data leakage from improper preprocessing sequences, (2) intentional vagueness in methodological details that prevents replication, (3) inadequate temporal validation for transaction data, and (4) metric manipulation through recall optimization at precision's expense. 
The findings of the last section suggest that methodological rigor matters more than algorithmic sophistication. To demonstrate this, we will present a deliberately flawed MLP implementation with SMOTE applied before train-test splitting - a clear violation of proper evaluation protocol. Despite this fundamental flaw, we anticipate this method will outperform many existing approaches, underscoring how data leakage can overshadow algorithmic advantages. This serves as a cautionary example that sophisticated techniques cannot compensate for basic methodological failures in fraud detection research.

The flawed methodology used in this investigation, to differentiate deceptive transactions from genuine ones, focuses on the most prevalent vice from the literature; balancing the data before splitting it into train/val/test partitions. It consists of the integration of two modules central to the detection of the fraudulent card transactions, \textit{viz.} a SMOTE module and a multilayer perceptron (MLP) network.
\subsection{Synthetic Minority Over-sampling Technique (SMOTE)}
To address the unbalanced data, we employ oversampling using the SMOTE, which is particularly effective in mitigating class imbalance issues in classification tasks. SMOTE generates synthetic samples of the minority class to balance the class distribution~\cite{chawla2002smote}.

The synthetic sample is generated using the following rule:
\begin{equation}
    x_{new} = x_{i} + \delta \times (x_{k} - x_{i}),
\end{equation}
where:\\
\hspace*{1cm}%
\begin{minipage}{.8\textwidth}%
$x_{new}$ is the new synthetic data point,
\\$x_{i}$ is the original minority class data point,
\\$x_{k}$ is one of the k nearest neighbors of $x_{i}$, and
\\ $\delta$ is a arbitrary value between zero and one.
\end{minipage}%

As far as the python implementation is concerned, the SMOTE related module is imported from the \texttt{imblearn} (\texttt{imbalanced-learn}) library\footnote{\url{https://imbalanced-learn.org/stable/references/generated/imblearn.over\_sampling.SMOTE.html}}. 
The \texttt{imblearn} package includes techniques for handling unbalanced datasets and is built on scikit-learn, an open-source Python library that provides simple and efficient tools for data mining and data analysis. 
This approach helps in creating a more balanced dataset, thereby improving the performance of classification models.
\subsection{The Multilayer Perceptron (MLP) module}
\begin{figure}[h!]
    \centering
    \includegraphics[width=4in]{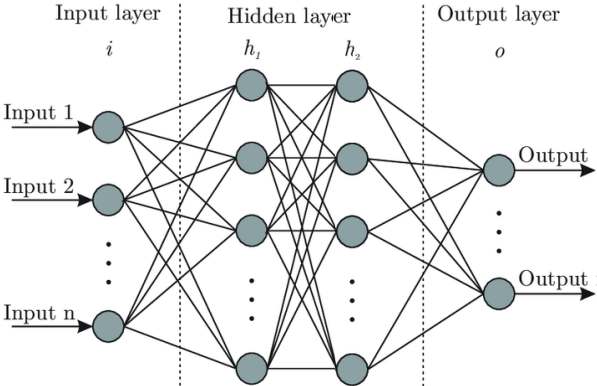}
    \caption{A generic MLP architecture}
    \label{fig:ANN}
\end{figure}

An Artificial Neural Network (ANN) is a computational model inspired by the structure and function of biological neural networks, designed to process inputs and learn patterns akin to human cognition. A typical MLP, a type of ANN (as illustrated in Fig.~\ref{fig:ANN}), consists of interconnected layers of nodes (neurons), where each connection is associated with a weight that is adjusted during training. Due to their ability to model complex, non-linear relationships, ANNs are widely employed in tasks such as pattern recognition, regression, and classification.

An MLP typically comprises three key components:
\begin{itemize}
\item \textbf{Input Layer:} Receives raw data and distributes it to the subsequent layer. Each neuron in this layer corresponds to a feature or attribute of the input data.
\item \textbf{Hidden Layers:} Perform computations and feature extraction, enabling the network to capture intricate patterns in the data.
\item \textbf{Output Layer:} Produces the final prediction or classification result. The number of neurons in this layer is determined by the specific task (e.g., one neuron for binary classification, multiple neurons for multi-class classification).
\end{itemize}

The fundamental building blocks of an ANN are neurons, which process inputs by applying an activation function to the weighted sum of their inputs and producing an output. This output is then propagated to subsequent neurons, facilitating the network's ability to learn and adapt~\cite{Amir-Al}.

% \begin{figure}[h!]
%     \centering
%     \includegraphics[width=4in]{logos/ANN_module.png}
%     \caption{The ANN module}
%     \label{fig:ANN_modle}
% \end{figure}
% % 
% To demonstrate the effectiveness of polynomial feature engineering, we are using a very simple ANN architecture. The core argument is that, with thorough feature engineering, there is no need for complex network architectures. As illustrated in Fig.~\ref{fig:ANN_modle}, our ANN module consists of only one hidden layer with 16 neurons, utilizing the rectified linear unit (ReLU) activation function. The input layer has 32 neurons, while the output layer has a single neuron with a sigmoid activation function, suitable for binary classification.
% 
\begin{figure}[h]
     \begin{python}
# SMOTE already applied to the whole dataset
# Followed by stratified train/test split
# the MLP
model = Sequential([
        # N: The number of neurons in the only hidden layer 
    Dense(N, activation='relu'),
    Dense(1, activation='sigmoid')
])
model.compile(optimizer='adam', loss='binary_crossentropy', metrics=['accuracy'])
\end{python}
    \caption{The Flawed MLP Model}
    \label{code:mlp_model}
\end{figure}
To emphasize on our argument, we are using a very simple MLP architecture. The core argument is that, with a casual leaky approach, there is no need of sophistication. Our MLP module consists of only one hidden layer with an arbitrary number of neurons ($N$), utilizing the rectified linear unit (ReLU) activation function. The input layer has $16$ (arbitrary) neurons, while the output layer has a single neuron with a sigmoid activation function, suitable for binary classification. Assuming that the data is subjected to SMOTE and then split to the training and testing parts, with \text{TensorFlow} environment, the model would have the code given in Figure~\ref{code:mlp_model}.

\subsection{Results and Analysis}
\begin{figure}[h!]
    \centering
    \subfloat[PRC]{\includegraphics[width=7cm]{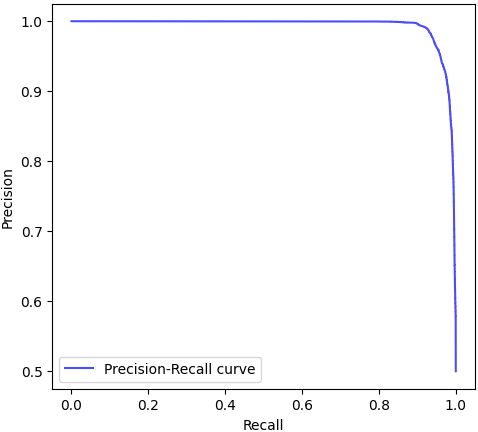}}\quad
    \subfloat[ROC]{\includegraphics[width=7cm]{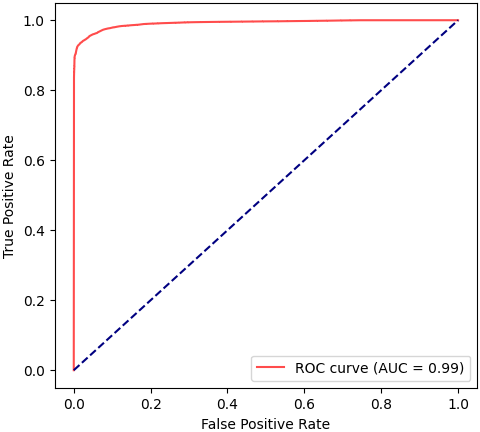}}\\
    \caption{Test results after applying the MLP with no hidden layer ($N=0$).}
    \label{fig:0N}
\end{figure}

To further demonstrate our argument about the disproportionate impact of methodological flaws versus algorithmic sophistication, we conducted an extreme simplification test: eliminating the hidden layer entirely ($N=0$) while maintaining the data leakage from SMOTE application before train-test splitting. Figure~\ref{fig:0N} presents the Precision-Recall Curve (PRC) and Receiver Operating Characteristic Curve (ROC) for this minimal configuration.
Despite this extreme simplification, Table~\ref{tab:my_label} shows we achieved remarkably high performance metrics: 94\% recall and 97.6\% precision. This demonstrates that:
\begin{itemize}
\item A single output neuron with data leakage can outperform many sophisticated models
\item The data leakage from improper SMOTE application provides more benefit than architectural complexity
\item Such results are artificially inflated and would not generalize to real-world scenarios
\end{itemize}
\begin{table}
    \centering
    \begin{tabular}{|c|c|c|c|c|c|}\hline
        No. & N & Accuracy & Precision & Recall & F1\\\hline
         1& 0 & 0.958 & 0.976 & 0.939 & 0.958\\\hline
         2& 1 &  0.959 & 0.985 & 0.932 & 0.957\\\hline
         3& 2 & 0.967 & 0.976 & 0.958 & 0.967\\\hline
         4& 4 & 0.982 & 0.980 & 0.983 & 0.982\\\hline
         5& 6 & 0.982 & 0.985 & 0.979 & 0.982\\\hline
         6& 8 & 0.986 & 0.988 & 0.985 & 0.986\\\hline
         7& 10 & 0.992 & 0.989 & 0.994 & 0.992\\\hline
         8& 12 & 0.992 & 0.991 & 0.992 & 0.992\\\hline
         9& 16 & 0.996 & 0.992 & 0.999 & 0.996\\\hline
    \end{tabular}
    \caption{Our MLP with various number of neurons (N) in its hidden layer}
    \label{tab:my_label}
\end{table}

\begin{figure}[!htbp]
    \centering
    \subfloat[$N=1$]{\includegraphics[width=5.5cm]{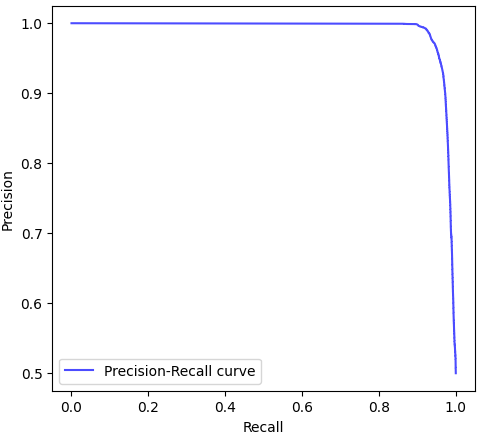}\quad\includegraphics[width=5.5cm]{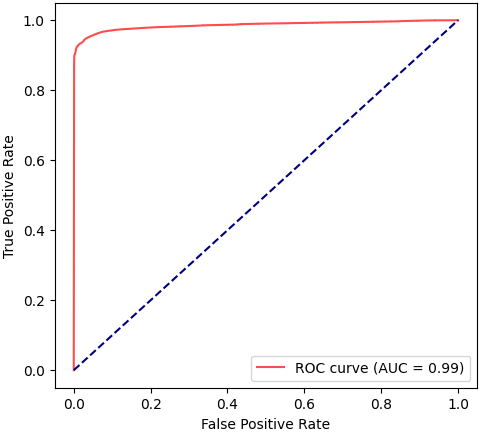}}\\
    \subfloat[$N=8$]{\includegraphics[width=5.5cm]{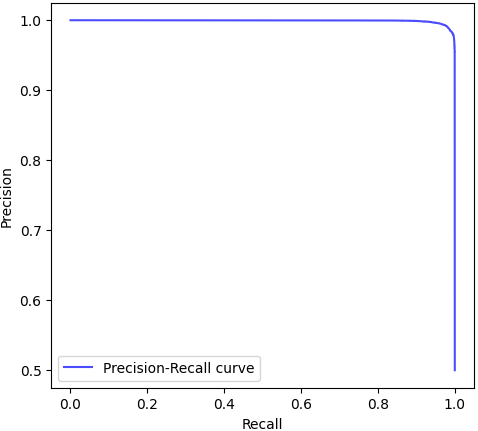}\quad\includegraphics[width=5.5cm]{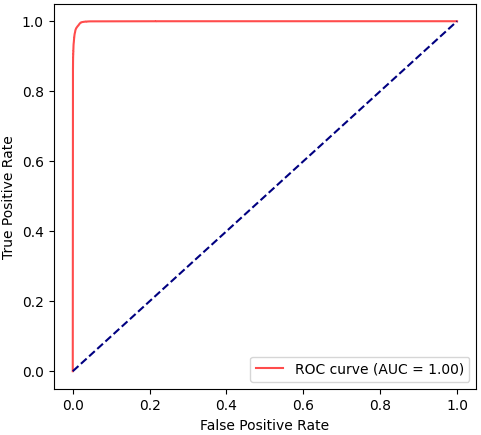}}\\
    \subfloat[$N=12$]{\includegraphics[width=5.5cm]{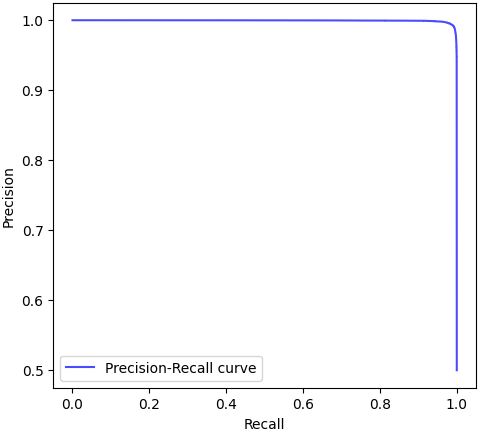}\quad\includegraphics[width=5.5cm]{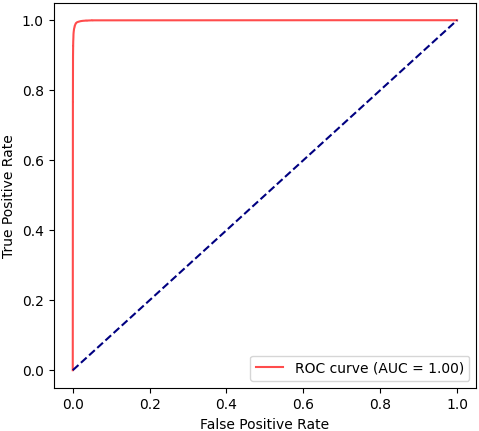}}\\
    \subfloat[$N=16$]{\includegraphics[width=5.5cm]{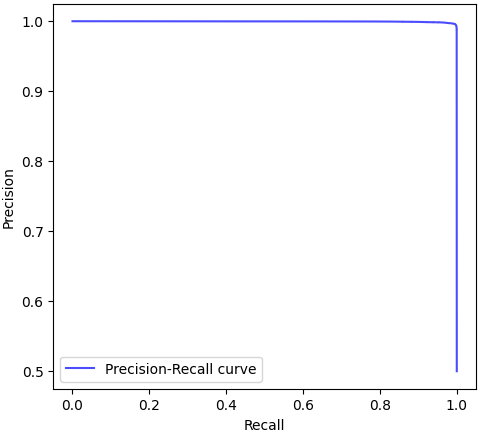}\quad\includegraphics[width=5.5cm]{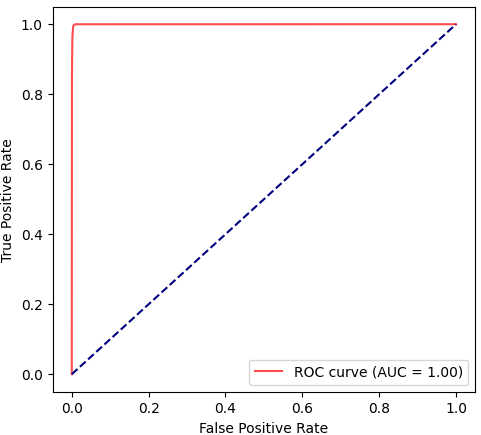}}\\
    \caption{Pairwise PRC(left) and ROC (right) test results of the MLP method with {$N$} neurons in a single hidden layer.}
    \label{fig:N}
\end{figure}

We then systematically increased the complexity by introducing a hidden layer and varying the number of neurons ($N=1$ to $N=16$). As shown in Table~\ref{tab:my_label}, performance metrics improved incrementally with more neurons, reaching near-perfect scores (99.9\% recall) at $N=16$. Figure~\ref{fig:N} illustrates the PRC and ROC curves for selected configurations.
% After introducing a simple hidden layer, we varied the number of neurons in the hidden layer, with maximum being $N=16$, while starting with a single hidden neuron ($N=1$). The results are illustrated for various values of $N$ in Table~\ref{tab:my_label}. One can readily observe that the already formidable results are further improving as we go along down the table, reaching $99\%$ by $N=10$. Figure~\ref{fig:N} plots the corresponding PRC and ROC for selected values of $N$. 
% 

Following are the important points to note:
\begin{itemize}
\item These results demonstrate how data leakage can overshadow architectural improvements.
\item Performance gains from adding neurons are marginal compared to the initial boost from data leakage.
\item The near-perfect metrics at $N=16$ (even beyond $N=8$) are statistically implausible for real-world fraud detection.
\item This experiment underscores that proper evaluation methodology matters more than model complexity.
\end{itemize}

This case study vividly illustrates how fundamental methodological flaws can produce deceptively impressive results that would fail in real-world deployment. The key takeaway is that the evaluation of rigor must precede architectural sophistication in fraud detection research.
% Such incredible results were, with the simplest of networks, the consequence of failure to recognize the importance of correct preprocessing. In this one example, we only dealt with the most prevalent flaw from the literature and showed that if you do not follow a proper sequence of preprocessing, you may end up with inflated results far from reality. 
\section{Conclusion}\label{sec_concl}
Our analysis reveals that methodological rigor is far more critical than algorithmic sophistication in fraud detection research. Through deliberate experimentation with flawed evaluation protocols, we demonstrated how even simple models can achieve deceptively impressive results when fundamental methodological principles are violated. The extreme case of a single-neuron model outperforming sophisticated architectures - solely due to data leakage from improper SMOTE application - serves as a stark illustration of how evaluation flaws can overshadow algorithmic advantages.

Beyond these technical considerations, our findings must be contextualized within the broader academic ecosystem. The peer review system faces significant challenges including reviewer fatigue, time constraints, and overwhelming submission volumes. These issues are exacerbated by publication pressures, financial incentives tied to rapid dissemination, and citation practices that may inadvertently inflate impact metrics. The current academic reward structure, which heavily weights publication quantity and citation counts for career advancement, can sometimes incentivize quantity over quality. While understandable given institutional pressures, this emphasis may occasionally compromise research rigor and originality.

These systemic factors contribute to a growing disconnect between academic research and industrial practice. As researchers, we often find ourselves playing catch-up to industry innovation, with many academic publications essentially formalizing concepts that have already gained practical traction. The review process itself may sometimes prioritize presentation quality over substantive contribution, allowing incremental or ambiguous work to pass through.

It is crucial to emphasize that this critique is general in nature and not directed at any specific work cited in this study. All references were selected based on their relevance and alignment with our discussion, and should not be construed as examples of the concerns raised. Rather, our analysis aims to highlight systemic issues that affect the field as a whole, with the goal of fostering more rigorous and impactful research practices.

Moving forward, we advocate for:

Stricter evaluation protocols that prioritize methodological soundness over novel architectures
Enhanced transparency in reporting preprocessing pipelines and evaluation methodologies
Better alignment between academic research and industrial needs
Reform of incentive structures to reward quality and impact over publication quantity
By addressing these challenges, we can bridge the gap between academic research and practical applications, ultimately advancing the field of fraud detection in more meaningful ways.

\end{document}